\documentclass[letterpaper]{article} 
\usepackage{aaai2027}  
\usepackage[hyphens]{url}  
\usepackage{graphicx} 
\usepackage{natbib}  
\usepackage{caption} 
\usepackage{algorithm}
\usepackage[noEnd=true,indLines=true]{algpseudocodex}
\usepackage{amsmath}
\usepackage{amssymb}
\usepackage{stmaryrd}

\newcommand{\suf}[1]{^{\mathrm{#1}}}
\newcommand{\sub}[1]{_{\mathrm{#1}}}
\newcommand{\funcname}[1]{\ensuremath{\mathsf{#1}}}
\newcommand{\defeq}{\mathrel{\mathop:}=}

\newcommand{\Input}[1]{%
  \item[\textbf{Input:}~#1]
}

\renewcommand{\Output}[1]{%
  \item[\textbf{Output:}~#1]
}

\newcommand{\Notation}[1]{%
  \item[\textbf{Notation:}~#1]
}

\newcommand{\Continue}{\textbf{continue}}

\providecommand{\cref}[1]{%
  Algorithm~\ref{#1}%
}

\newcommand{\C}{\ensuremath{\mathcal{C}}}
\renewcommand{\S}{\ensuremath{\mathcal{S}}}

\newcommand{\N}{\ensuremath{\mathcal{N}}}
\newcommand{\Q}{\ensuremath{\mathcal{Q}}}

\newcommand{\from}{\suf{from}}
\renewcommand{\to}{\suf{to}}
\newcommand{\goal}{\suf{goal}}
\newcommand{\init}{\suf{init}}
\newcommand{\new}{\suf{new}}

\newcommand{\lacamstar}{%
  \ensuremath{\mathrm{LaCAM}^{\ast}}%
}
\newcommand{\PIBT}{\funcname{PIBT}}
\newcommand{\neigh}{\funcname{neigh}}
\newcommand{\dist}{\funcname{dist}}
\newcommand{\cost}{\funcname{cost}}
\newcommand{\interrupt}{\funcname{interrupt}}
\newcommand{\backtrack}{\funcname{backtrack}}
\newcommand{\h}{\funcname{h}}

\newcommand{\open}{\ensuremath{\mathit{Open}}}
\newcommand{\explored}{\ensuremath{\mathit{Explored}}}
\newcommand{\config}{\ensuremath{\mathit{config}}}
\newcommand{\tree}{\ensuremath{\mathit{tree}}}
\newcommand{\parent}{\ensuremath{\mathit{parent}}}
\newcommand{\neighbors}{\ensuremath{\mathit{neigh}}}

\newcommand{\nosolution}{%
  \ensuremath{\mathtt{NO\_SOLUTION}}%
}
\newcommand{\failure}{%
  \ensuremath{\mathtt{FAILURE}}%
}
\newcommand{\valid}{%
  \ensuremath{\mathtt{VALID}}%
}
\newcommand{\invalid}{%
  \ensuremath{\mathtt{INVALID}}%
}

\usepackage{newfloat}
\usepackage{listings}

\DeclareCaptionStyle{ruled}{
  labelfont=normalfont,
  labelsep=colon,
  strut=off
} 

\floatstyle{ruled}
\newfloat{listing}{tb}{lst}{}
\floatname{listing}{Listing}

\title{ITA-LaCAM: A Complete and Scalable TAPF Solver\\via Assignment-Aware Configuration-Space Search}
\author {
    Yimin Tang\textsuperscript{\rm 1,3},
    Han Zhang\textsuperscript{\rm 3},
    Shao-Hung Chan\textsuperscript{\rm 3},
    Junsoo Kim\textsuperscript{\rm 1},
    Erdem B{\i}y{\i}k\textsuperscript{\rm 1},
    Sven Koenig\textsuperscript{\rm 2},
    Jingkai Chen\textsuperscript{\rm 3}
}
\affiliations {
    \textsuperscript{\rm 1}Thomas Lord Department of Computer Science, University of Southern California\\
    \textsuperscript{\rm 2}Department of Computer Science, University of California Irvine\\
    \textsuperscript{\rm 3}Symbotic\\
    \{yimintan,junsooki,biyik\}@usc.edu, svenk@uci.edu, \{hanzhang,shchan,jichen\}@symbotic.com
}

\begin{document}

\maketitle


\begin{abstract}
Combined Target Assignment and Path Finding (TAPF) requires assigning targets for agents while simultaneously planning collision-free paths. We present ITA-LaCAM, a complete and scalable TAPF solver inspired by LaCAM and ITA-CBS. In ITA-LaCAM, each joint-configuration node carries an agent-to-target matching. When a successor is generated, ITA-LaCAM incrementally repairs the matching for the agents that moved and uses the targets to guide PIBT successor generation. This design enables adaptive reassignment without explicitly enumerating the combinatorial assignment space, while preserving LaCAM's completeness and scalability. Across 9,760 benchmark instances on eight maps with 5--200 agents, ITA-LaCAM solved 100\% of the instances, compared with 95.6\% for IR-TAPF configured with DBS-Hungarian. ITA-LaCAM found an initial solution faster in 84.0\% of the comparisons and achieved a lower sum of costs in 65.0\% of the instances solved by both methods.
\end{abstract}



\section{Introduction}

Multi-Agent Path Finding (MAPF) plans collision-free paths for multiple agents from their starts to distinct goals while optimizing a cost criterion. It models coordination in warehouse automation, transportation, and aerial swarms. Since optimal MAPF is NP-hard~\cite{yu2013structure,stern2019multi}, optimal solvers such as Conflict-Based Search (CBS)~\cite{sharon2015conflict} and $M^*$~\cite{wagner2011m} face scalability challenges as agent count and density increase. Faster suboptimal methods include Prioritized Planning (PP)~\cite{erdmann1987multiple,silver2005cooperative}, PBS~\cite{ma2019searching}, LaCAM~\cite{okumura2023lacam}, MAPF-LNS2~\cite{li2022mapf}, and their variants~\cite{chan2023greedy,okumura2024engineering}. These methods find solutions much faster on large or congested instances but generally lack optimality guarantees.

In many applications, targets are not fixed in advance. Warehouse robots, for example, may choose among shelves, storage locations, or workstations~\cite{wurman2008coordinating}. Combined Target-Assignment and Path-Finding (TAPF)~\cite{cbm2016} assigns each agent one allowable target while planning collision-free paths to all assigned targets. MAPF is the special case with one allowable target per agent. TAPF thus inherits MAPF's difficulty while adding the tightly coupled assignment decision; a poor assignment can create congestion and greatly complicate path finding. Optimal TAPF algorithms include CBM~\cite{cbm2016}, CBS-TA~\cite{honig2018conflict}, and ITA-CBS~\cite{tang2023solving}. Although they provide strong solution-quality guarantees, like CBS-based MAPF solvers, their conflict-resolution trees become expensive on large, dense instances. Bounded-suboptimal variants such as ECBS-TA~\cite{honig2018conflict} and ITA-ECBS~\cite{tang2024ita} improve scalability with bounded costs but retain much of the CBS-style search overhead.

\providecommand{\mapfigdir}{./figures}
\begin{figure}[t]
    \centering
    \begin{tabular}{@{}ccc@{}}
        \includegraphics[width=0.295\columnwidth]{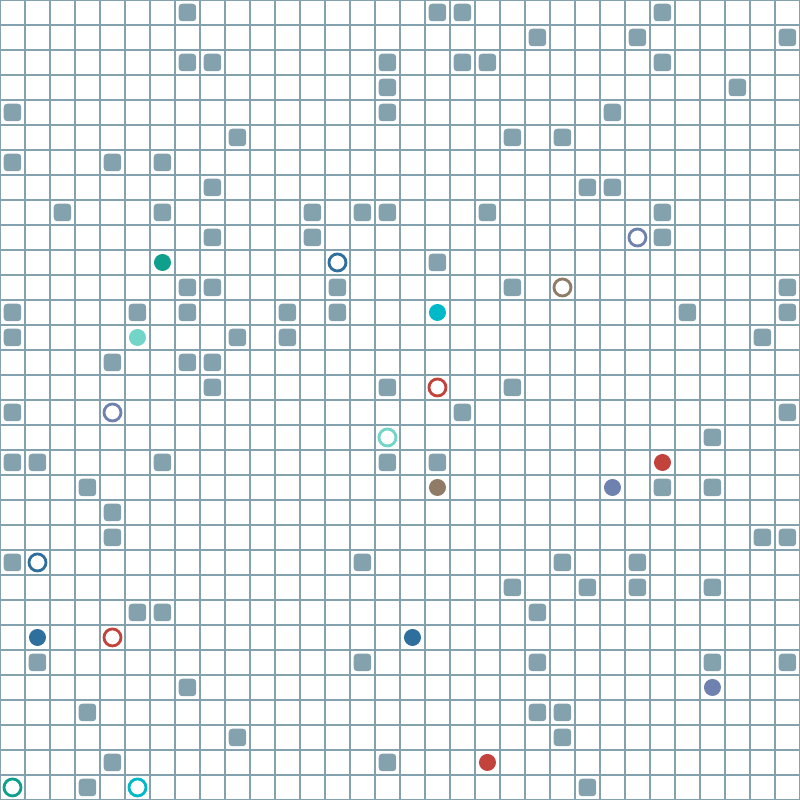} &
        \includegraphics[width=0.295\columnwidth]{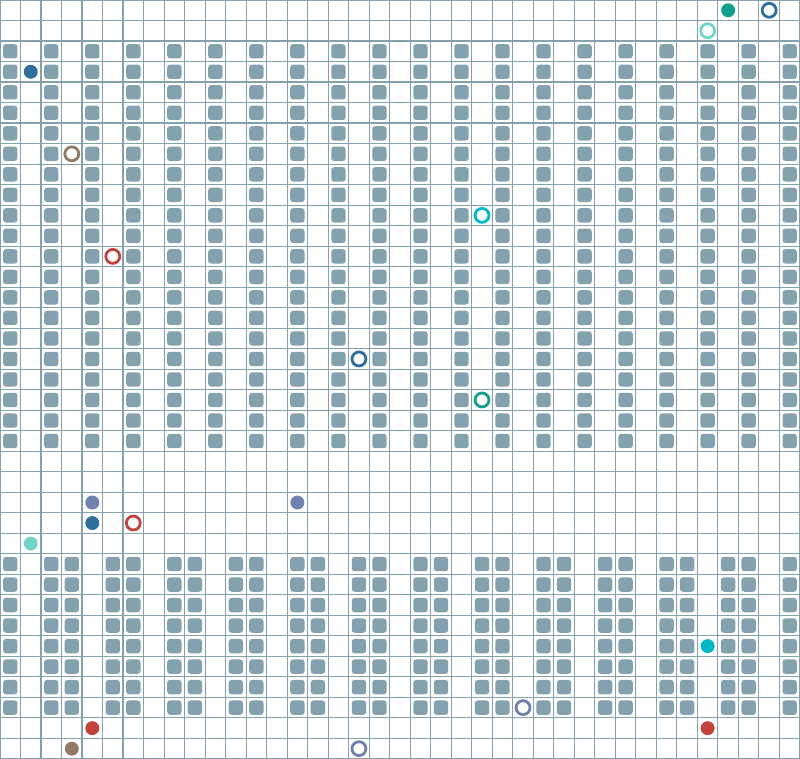} &
        \includegraphics[width=0.295\columnwidth]{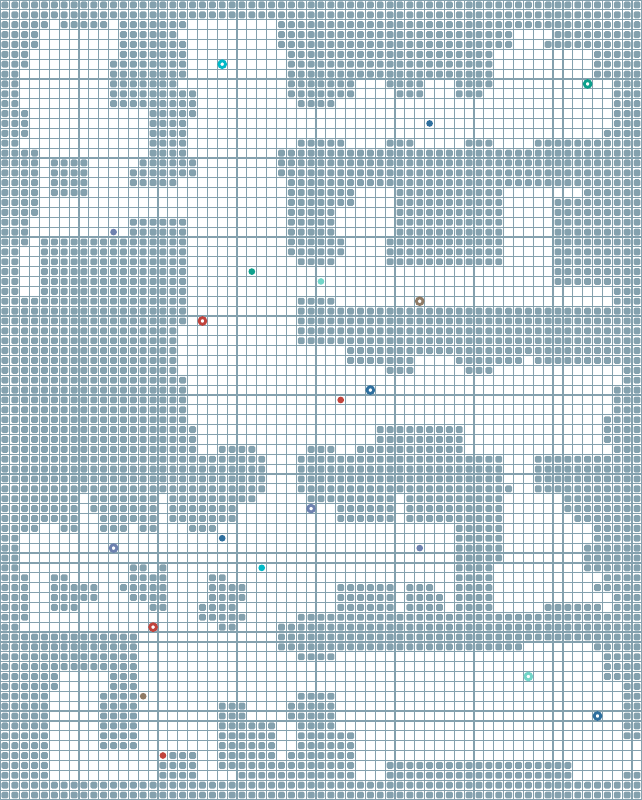} \\
        {\small\shortstack{(a) Random\\$32\times32$}} &
        {\small\shortstack{(b) Symbotic\\$39\times37$}} &
        {\small\shortstack{(c) den312d\\$65\times81$}} \\[2pt]
        \includegraphics[width=0.295\columnwidth]{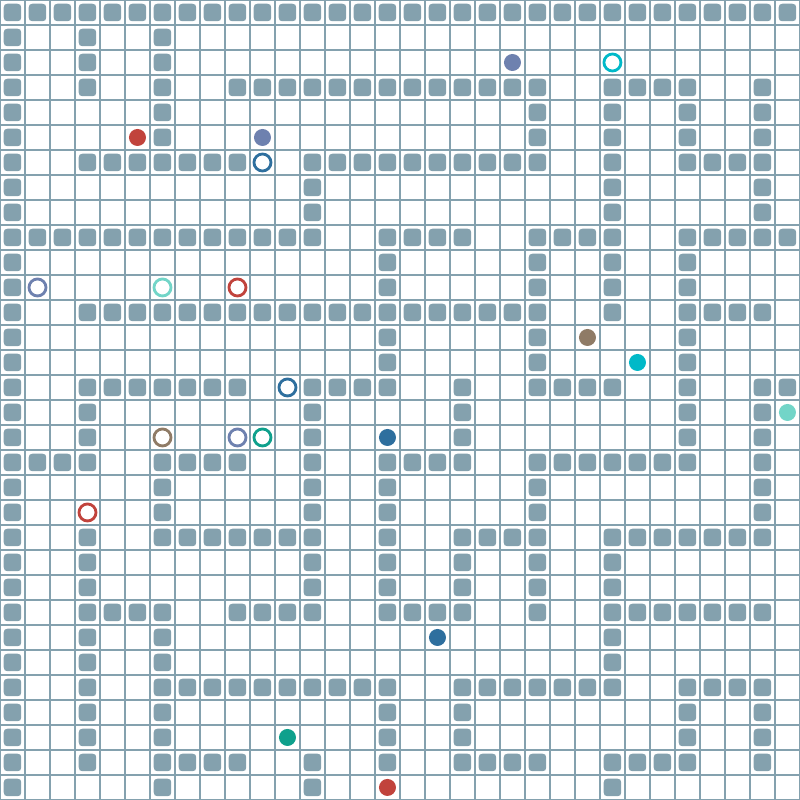} &
        \includegraphics[width=0.295\columnwidth]{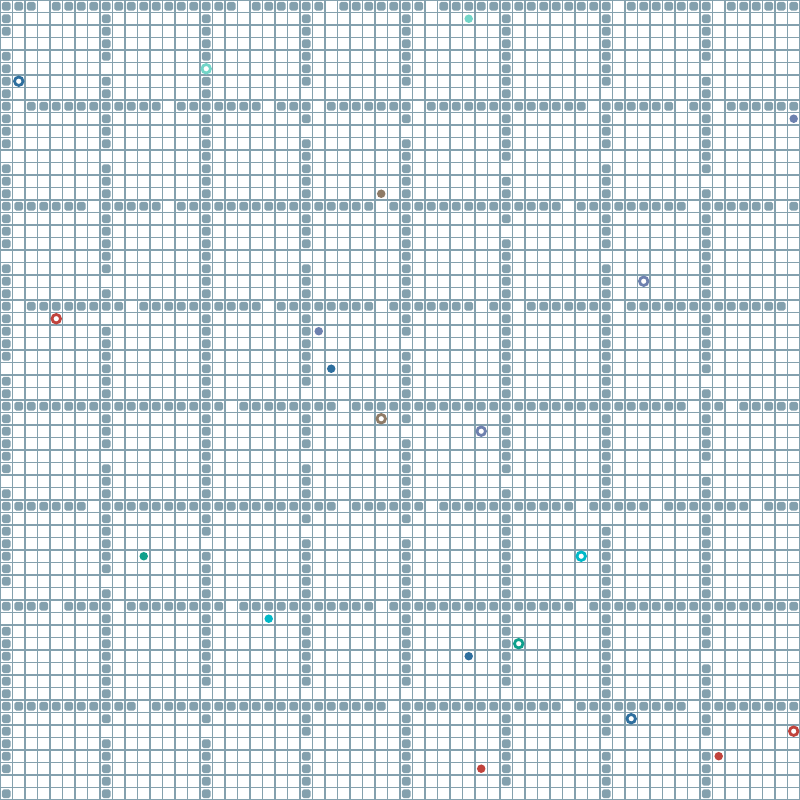} &
        \includegraphics[width=0.295\columnwidth]{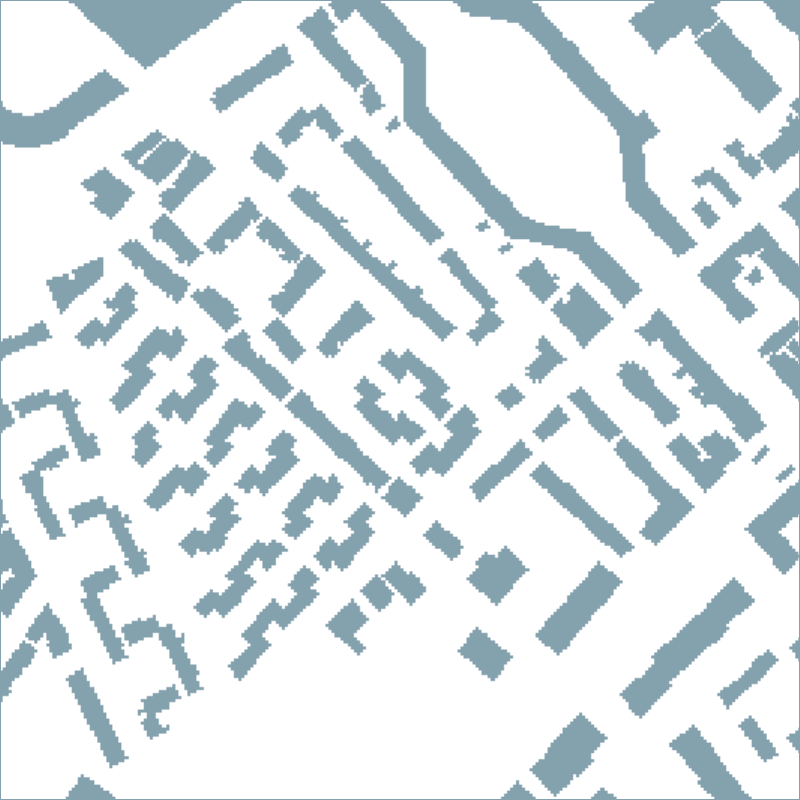} \\
        {\small\shortstack{(d) Maze\\$32\times32$}} &
        {\small\shortstack{(e) Room\\$64\times64$}} &
        {\small\shortstack{(f) Boston\\$256\times256$}} \\[2pt]
        \multicolumn{3}{c}{%
            \begin{tabular}{@{}cc@{}}
                \includegraphics[width=0.45\columnwidth]{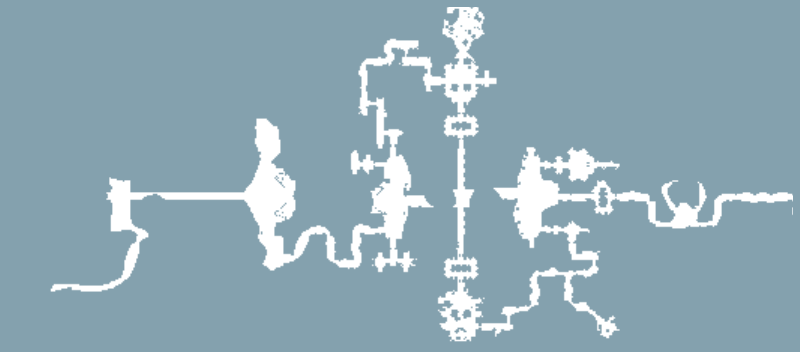} &
                \includegraphics[width=0.45\columnwidth]{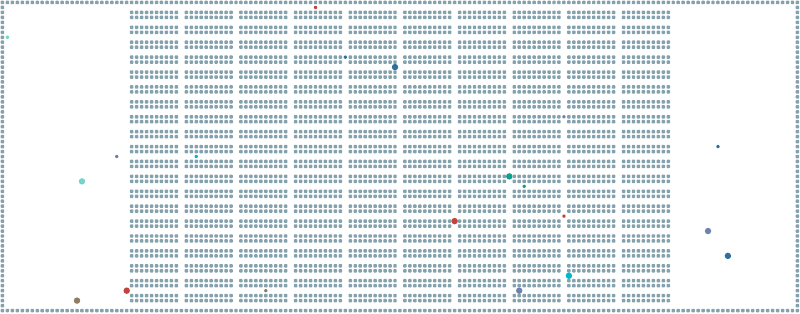} \\
                {\small\shortstack{(g) orz900d\\$1491\times656$}} &
                {\small\shortstack{(h) Warehouse\\$161\times63$}}
            \end{tabular}}
    \end{tabular}
    \caption{The eight benchmark maps used in our experiments.}
    \label{fig:benchmark-maps}
\end{figure}

IR-TAPF~\cite{kumagai2026alternating} improves scalability over optimal and bounded-suboptimal TAPF solvers through iterative refinement of target assignments and paths. Like CBS-TA, it follows an assignment-then-planning decomposition: it fixes an assignment and solves the induced MAPF instance using LaCAM. Path feedback can update the assignment only after a MAPF solution, so an unfavorable assignment may delay the first solution and leave fewer refinement iterations within the time limit.

To shorten this feedback loop, we propose ITA-LaCAM, inspired by ITA-CBS and built on \lacamstar{}~\cite{okumura2023improvinglacamscalableeventually}. \lacamstar{} is a complete MAPF solver that searches joint configurations and uses Priority Inheritance with Backtracking (PIBT)~\cite{okumura2022priority} to generate successors lazily. ITA-LaCAM augments each high-level node with a minimum-distance matching between agents and allowable targets, without enumerating the combinatorial assignment space. This matching defines temporary PIBT targets, the heuristic, and an assignment-dependent search cost. Each successor incrementally repairs the matching for the agents that moved, updating target assignment online while retaining the underlying configuration search. Our contributions are as follows:
\begin{itemize}
    \item We introduce ITA-LaCAM, a fast and complete solver for TAPF that integrates target assignment directly into configuration search.
    \item We show empirically that ITA-LaCAM finds first solutions faster than IR-TAPF in 84.0\% of cases and lower SOC in 65.0\%, while 90\% of its paired solutions are within 10.2\% of the ITA-CBS optimum.
\end{itemize}


\section{Problem Definition}

The Combined Target-Assignment and Path-Finding (TAPF) problem is defined on an undirected graph $G=(V,E)$. Each vertex represents a possible agent location, and each unit-cost edge permits movement between two locations. Self-loops represent wait actions, so every move or wait takes one timestep. Let $I=\{1,\ldots,N\}$ be the agents, where agent $i$ starts at $s_i\in V$, and let $\mathcal{G}=\{g_1,\ldots,g_M\}\subseteq V$ be the targets, with $M\geq N$. 

Eligibility is encoded by a binary \emph{target matrix} $A\in\{0,1\}^{N\times M}$, whose rows correspond to agents and columns to targets. Entry $A[i][j]=1$ iff agent $i$ may be assigned to $g_j$. Agent $i$'s target set is therefore $\mathcal{G}_i=\{g_j\in\mathcal{G}\mid A[i][j]=1\}$. A target assignment is a function $\tau:I\rightarrow\mathcal{G}$. It is feasible if $\tau(i)\in\mathcal{G}_i$ for every agent and injective, so different agents receive different targets. Such an assignment exists only if the target matrix admits a matching that covers all agents. This injectivity is necessary because agents remain at their targets after arrival. Unlike MAPF, the assignment is part of the solution rather than fixed in the input.

Let $v_t^i\in V$ be agent $i$'s location at timestep $t$, and let $\pi_i=[v_0^i,\ldots,v_{T^i}^i]$ be its path from $s_i$ to $\tau(i)$. Consecutive locations must satisfy $(v_t^i,v_{t+1}^i)\in E$, including self-loops for wait actions. After arrival, the agent remains at its target, i.e., $v_t^i=v_{T^i}^i$ for all $t>T^i$. Thus, $T^i$ is the arrival time and the cost of $\pi_i$; it counts all move and wait actions before arrival but not waiting afterward. For distinct agents $i$ and $j$, a \emph{vertex collision} occurs at timestep $t$ if $v_t^i=v_t^j$. An \emph{edge collision} occurs between timesteps $t$ and $t+1$ if $v_t^i=v_{t+1}^j$ and $v_{t+1}^i=v_t^j$. These conditions apply at every timestep, including after one agent has reached its target. A set of paths is collision-free if neither type occurs.

A TAPF solution consists of a feasible assignment $\tau$ and paths $\{\pi_i\mid i\in I\}$ satisfying:
\begin{enumerate}
\item $v_0^i=s_i$ and $v_{T^i}^i=\tau(i)$ for every agent $i$;
\item every action is valid, i.e., $(v_t^i,v_{t+1}^i)\in E$ for all $0\leq t<T^i$; and
\item the paths are collision-free.
\end{enumerate}

We minimize the sum of costs (SOC), also called flowtime: $\mathrm{SOC}=\sum_{i=1}^{N}T^i$.

\section{Related Work}

\subsection{Multi-Agent Path Finding}

MAPF asks for collision-free paths from agents' start locations to pre-assigned targets while minimizing a cost function. Both search- and learning-based methods have been developed~\cite{silver2005cooperative,luna2011push,wang2008fast,standley2010finding,standley2011complete,sharon2015conflict,wagner2015subdimensional,wagner2011m,ma2019searching}. Learning-based approaches either use trained policies to construct solutions directly~\cite{sartoretti2019primal,andreychuk2024mapf,jiang2025deploying,tang2025railgununifiedconvolutionalpolicy} or use neural models to guide search~\cite{huang2022anytime,veerapaneni2024improvinglearntlocalmapf,agaskar2026deepfleetmultiagentfoundationmodels,jain2025graphattentionguidedsearchdense}.

Among search-based methods, Conflict-Based Search (CBS)~\cite{sharon2015conflict} is a widely used optimal algorithm. It plans individual paths and resolves collisions by adding constraints and replanning affected agents. Bounded-suboptimal variants such as ECBS~\cite{barer2014suboptimal} and EECBS~\cite{li2021eecbs} improve scalability by relaxing optimality. Prioritized Planning (PP)~\cite{erdmann1987multiple,silver2005cooperative} instead plans agents sequentially, requiring each agent to avoid earlier paths. This greatly reduces search but sacrifices completeness.

More directly related to our work, LaCAM and its variants~\cite{okumura2023lacam,okumura2024engineering} provide a scalable suboptimal alternative. LaCAM searches joint configurations and uses PIBT to generate valid successors lazily, avoiding explicit enumeration of the joint action space. ITA-LaCAM extends this configuration-search framework from MAPF to TAPF.


\subsection{Combined Target-Assignment and Path-Finding}

Unlike MAPF, TAPF must determine both an agent-to-target assignment and collision-free paths. A suitable assignment can substantially reduce the difficulty of the induced path-finding problem. Existing TAPF algorithms include CBM~\cite{cbm2016}, CBS-TA and ECBS-TA~\cite{honig2018conflict}, ITA-CBS~\cite{tang2023solving}, and ITA-ECBS~\cite{tang2024ita}. CBM combines CBS with a flow-based low-level solver to minimize makespan optimally. CBS-TA follows an assignment-then-planning decomposition and solves the MAPF problem induced by each target assignment using CBS. ITA-CBS instead integrates target assignment into a single CBS constraint tree and updates the assignment as constraints are added. ECBS-TA and ITA-ECBS are their bounded-suboptimal counterparts. Although these methods provide optimality or bounded-suboptimality guarantees, their CBS-based search limits scalability on large instances.

IR-TAPF~\cite{kumagai2026alternating} is a scalable suboptimal TAPF method based on iterative refinement. It first computes a target assignment and uses LaCAM to solve the induced MAPF instance. It then analyzes the complete path solution, updates the assignment, and repeats this process within the time limit while retaining the best solution found. This design improves scalability, but assignment updates are available only after a complete MAPF solution has been generated. ITA-LaCAM follows the integration idea of ITA-CBS at a finer granularity: it updates the target assignment at each newly generated configuration, shortening the feedback loop between assignment and path finding.


\section{Method}

This section presents ITA-LaCAM, which extends \lacamstar{} from MAPF to TAPF by integrating target assignment into configuration search. We first motivate this design from the configuration-space view, then describe its incremental Hungarian updates and PIBT-with-swap successor generation.

\subsection{Configuration-Space Motivation}

We begin with \lacamstar{}, summarized in Algorithm~\ref{algo:star}; without the bold operations, it is the original MAPF algorithm. \lacamstar{} performs an A*-like search over joint configurations $\Q=(v_1,\ldots,v_N)$, where $v_i\in V$ is agent $i$'s location. The initial configuration is $\S=(s_1,\ldots,s_N)$, and the fixed MAPF goal is $\Q_{\mathrm{goal}}=(g_1,\ldots,g_N)$.

A joint configuration $\Q$ is valid if every agent occupies a traversable vertex and no two agents occupy the same vertex, i.e., $v_i\neq v_j$ for all $i\neq j$. Two valid configurations $\Q=(v_1,\ldots,v_N)$ and $\Q'=(v'_1,\ldots,v'_N)$ are connected by a valid transition if every agent either waits in place or moves to an adjacent vertex, i.e., $v'_i=v_i$ or $(v_i,v'_i)\in E$, and no pair of agents swaps locations, i.e., there do not exist agents $i\neq j$ such that $v_i=v'_j$ and $v_j=v'_i$.

These configurations and transitions induce a valid configuration graph whose edges represent collision-free simultaneous actions. Every collision-free MAPF solution corresponds to a path from $\S$ to $\Q_{\mathrm{goal}}$ in this graph, and every such path is a collision-free solution. This configuration-space perspective supports different planning problems by changing the goal set, transition costs, or task information associated with each state. MAPF has one pre-assigned target per agent, whereas TAPF admits multiple goal configurations because its assignment is not fixed in advance. MAPD and lifelong MAPF can similarly use the same configuration space while successively reaching multiple goals.

We next consider how to explore this graph. CBS and $M^*$ first plan individual paths and resolve interactions after detecting collisions. Their intermediate states may therefore combine paths containing vertex or edge collisions and need not correspond to valid paths in the configuration graph. They reason in a relaxed space containing both the valid graph and additional invalid configurations or transitions. This larger space can be costly in large, congested instances where many conflicts must be detected and resolved. Prioritized Planning and PBS take the opposite approach. They impose priority relations and plan lower-priority agents while treating higher-priority agents as moving obstacles, substantially reducing the search space. However, this restricts search to a subset of the valid graph. A solution may require interactions or precedence relations inconsistent with the selected ordering, so these methods can fail even when the full graph contains a valid path. Applying A* to joint configurations searches exactly the valid graph but suffers from the curse of dimensionality. If each of $N$ agents has five actions, including waiting, one configuration may have up to $5^N$ successors. Explicitly generating and evaluating all such joint actions at every node is therefore impractical.

\lacamstar{} addresses this problem through lazy successor generation. Each high-level node maintains a low-level constraint tree that partially specifies agents' next locations. For each constraint, PIBT constructs one promising collision-free successor that satisfies it, while alternative constraints remain available for later visits. Progressive expansion eventually generates every valid successor without enumerating them all at once, so \lacamstar{} searches the full reachable graph while avoiding the upfront branching of direct A*.

The key observation is that feasible future transitions depend only on the current configuration and problem definition, not on how the configuration was reached. Given an appropriate heuristic toward a goal, \lacamstar{} can retain the same high-level search framework. Extending it to TAPF therefore mainly requires replacing MAPF's fixed-target guidance with assignment-aware guidance that selects temporary targets at each configuration and directs successor generation toward a feasible TAPF goal. This observation leads to ITA-LaCAM.

\subsection{Integrating Target Assignment into Configuration Search}

Extending \lacamstar{} to TAPF does not require changing its configuration-space search; the main challenge is to guide each configuration toward one of several feasible TAPF goals by determining which target each agent should currently pursue. A straightforward approach first computes an assignment and then solves the induced MAPF instance. CBS-TA follows this structure, while IR-TAPF updates each fixed assignment only after obtaining a complete MAPF solution. Both create a relatively long feedback loop between assignment and path finding: an unfavorable assignment may consume a complete MAPF call before path feedback becomes available.

Inspired by ITA-CBS, ITA-LaCAM computes a feasible minimum-distance matching $\N.\tau$ at every high-level node $\N$. Its temporary targets guide PIBT successor generation and the heuristic. Each new configuration immediately updates the matching from the new agent locations, allowing assignment and path finding to exchange feedback at every search node without changing \lacamstar{}'s underlying search. The assignment also defines the internal search cost, but does not permanently prune any valid successor before the first solution. Because every valid successor remains reachable through lazy expansion, ITA-LaCAM preserves \lacamstar{}'s completeness.

\subsection{ITA-LaCAM}

Algorithm~\ref{algo:star} presents the high-level search of ITA-LaCAM. The algorithm retains the configuration-search structure of \lacamstar{} while adding a target assignment $\N.\tau$ to every high-level node. The TAPF-specific operations are highlighted in bold.

\renewcommand{\open}{\ensuremath{\mathsf{OPEN}}}
\renewcommand{\explored}{\ensuremath{\mathsf{EXPLORED}}}
\newcommand{\dopen}{\ensuremath{\mathsf{DOPEN}}}
\newcommand{\fvalue}{\funcname{f}}

{
  \begin{algorithm}[t!]
    \caption{ITA-LaCAM.
      $\C\sub{init}$ denotes ``no constraint.''}
    \label{algo:star}

    \small
    \algrenewcommand\algorithmicif{\textnormal{if}}
    \algrenewcommand\algorithmicthen{\textnormal{then}}
    \algrenewcommand\algorithmicelse{\textnormal{else}}
    \algrenewcommand\algorithmicwhile{\textnormal{while}}
    \algrenewcommand\algorithmicdo{\textnormal{do}}
    \algrenewcommand\algorithmicfor{\textnormal{for}}
    \algrenewcommand\algorithmicend{\textnormal{end}}
    \algrenewcommand\algorithmicreturn{\textnormal{return}}
    \renewcommand{\Continue}{\textnormal{continue}}

    \begin{algorithmic}[1]

      \Input{\textbf{TAPF} instance, transition cost $\cost_e$, and heuristic $\h$}
      \Output{Solution, \nosolution, or \failure}
      \Notation{$\fvalue(\N)\defeq\N.g+\h(\N)$; $\spadesuit\defeq(\N\goal\neq\bot)$}

      \State Initialize $\open$ and $\explored$; $\N\goal\gets\bot$
      \label{algo:star:init-var}

      \State Initialize the root node $\N\init$
      \Statex \hspace{\algorithmicindent} $\N\init.\config\gets\S$, $\N\init.\tree\gets\llbracket\C\init\rrbracket$
      \Statex \hspace{\algorithmicindent} $\boldsymbol{\N\init.\tau\gets\funcname{task\_assignment}(\S)}$
      \Statex \hspace{\algorithmicindent} $\N\init.\parent\gets\bot$, $\N\init.\neighbors\gets\emptyset$
      \Statex \hspace{\algorithmicindent} $\N\init.g\gets 0$
      \label{algo:star:init-node}

      \State $\open.\funcname{push}(\N\init)$; $\explored[\S]\gets\N\init$

      \While{$\open\neq\emptyset\land\neg\interrupt()$}
        \label{algo:star:while-start}

        \State $\N\gets\open.\funcname{top}()$

        \If{$\boldsymbol{\funcname{is\_tapf\_goal}(\N.\config)}
          \land(\neg\spadesuit\lor\N.g<\N\goal.g)$}
          \State $\N\goal\gets\N$
        \EndIf
        \label{algo:star:goal}

        \If{$\spadesuit\land\fvalue(\N\goal)\leq\fvalue(\N)$}
          \State $\open.\funcname{pop}()$
          \State \Continue
        \EndIf
        \label{algo:star:branching}

        \If{$\N.\tree=\emptyset$}
          \State $\open.\funcname{pop}()$
          \State \Continue
        \EndIf

        \State $\C\gets\N.\tree.\funcname{pop}()$

        \State $\funcname{low\_level\_expansion}(\N,\C)$
        \label{algo:star:low-level-expansion}

        \State $\Q\new\gets\funcname{configuration\_generator}(\N,\C,\boldsymbol{\N.\tau})$
        \label{algo:star:configuration-generation}

        \If{$\Q\new=\bot$}
          \label{algo:star:generation-failure}
          \State \Continue
        \EndIf

        \If{$\explored[\Q\new]\neq\bot$}
          \label{algo:star:already-known}

          \State $\N.\neighbors.\funcname{append}\bigl(\explored[\Q\new]\bigr)$
          \label{algo:star:add-neighbor-known}

          \State $\dopen\gets\llbracket\N\rrbracket$
          \label{algo:star:propagation-start}

          \While{$\dopen\neq\emptyset$}

            \State $\N\from\gets\dopen.\funcname{pop}()$

            \For{$\N\to\in\N\from.\neighbors$}

              \State $g\gets\N\from.g+\cost_e(\N\from,\N\to)$

              \If{$g<\N\to.g$}
                \label{algo:star:g-value-pruning}

                \State $\N\to.g\gets g$
                \State $\N\to.\parent\gets\N\from$
                \State $\dopen.\funcname{push}(\N\to)$
                \label{algo:star:propagation-end}

                \If{$\spadesuit\land\fvalue(\N\to)<\fvalue(\N\goal)$}
                  \State $\open.\funcname{push}(\N\to)$
                  \label{algo:star:reinsert}
                \EndIf

              \EndIf
            \EndFor
          \EndWhile

        \Else

          \State Initialize a new node $\N\new$
          \Statex \hspace{\algorithmicindent} $\N\new.\config\gets\Q\new$, $\N\new.\tree\gets\llbracket\C\init\rrbracket$
          \Statex \hspace{\algorithmicindent} $\boldsymbol{\N\new.\tau\gets\funcname{task\_assignment}(\Q\new)}$
          \Statex \hspace{\algorithmicindent} $\N\new.\parent\gets\N$, $\N\new.\neighbors\gets\emptyset$
          \Statex \hspace{\algorithmicindent} $\N\new.g\gets\N.g+\cost_e(\N,\N\new)$
          \label{algo:star:new-node}

          \State $\open.\funcname{push}(\N\new)$
          \State $\explored[\Q\new]\gets\N\new$

          \State $\N.\neighbors.\funcname{append}(\N\new)$
          \label{algo:star:append-neighbor}

        \EndIf
      \EndWhile
      \label{algo:star:while-end}

      \If{$\spadesuit\land\open=\emptyset$}

        \State \Return $\backtrack(\N\goal)$
        \label{algo:star:best-incumbent-exhausted}

      \ElsIf{$\spadesuit$}

        \State \Return $\backtrack(\N\goal)$
        \label{algo:star:best-incumbent-interrupted}

      \ElsIf{$\open=\emptyset$}

        \State \Return \nosolution

      \Else

        \State \Return \failure

      \EndIf

    \end{algorithmic}
  \end{algorithm}
}

In Algorithm~\ref{algo:star}, a high-level node $\N$ represents a configuration $\N.\config$. The stack $\open$ contains nodes with successors left to generate, while $\explored$ maps each discovered configuration to its unique node. Each node also stores a parent for reconstructing the current path, a cost-to-come $\N.g$, and outgoing neighbors $\N.\neighbors$. The heuristic gives $\fvalue(\N)=\N.g+\h(\N)$. Because assignments can change between configurations, $\N.g$ is an assignment-dependent search cost that may differ from final-path SOC; incumbent comparisons use this internal cost. Here, $\bot$ denotes no valid value, $\C\init$ is an empty constraint, $\llbracket x\rrbracket$ initializes a container with $x$, and $\spadesuit$ indicates that an incumbent goal node $\N\goal$ exists.

The low-level constraint tree $\N.\tree$ generates successors one at a time. A constraint $\C$ fixes the next locations of zero or more agents. $\funcname{low\_level\_expansion}$ does not move the agents; it selects one unspecified agent and adds a child constraint for each move or wait action, retaining these alternatives for later visits. $\funcname{configuration\_generator}$ performs the complementary operation: it respects the fixed moves and uses PIBT, guided by $\N.\tau$, to move the remaining agents. It returns a collision-free $\Q\new$, or $\bot$ if the constraint cannot be completed. For example, consider agents at $(1,1)$ and $(1,3)$ with temporary targets $(2,1)$ and $(2,3)$. The empty constraint may first move both agents downward and generate $((2,1),(2,3))$. Meanwhile, the low-level expansion stores an alternative that makes agent~1 wait; when processed later, PIBT may generate $((1,1),(2,3))$. LaCAM thus tries a promising joint move without losing alternatives or enumerating every move combination. Returning $\bot$ rejects only the current constraint; $\N$ remains in $\open$ until its constraint tree is exhausted.

Every generated transition is recorded as a directed edge in $\N.\neighbors$. A new configuration creates a node, whereas a known configuration connects to its existing node in $\explored$. Since the new edge may provide a shorter path, Lines~\ref{algo:star:propagation-start}--\ref{algo:star:propagation-end} repeatedly propagate improved $g$-values and parent pointers with a queue.

Each high-level node stores an assignment $\N.\tau$ computed by the Hungarian algorithm. For configuration $\Q$, the cost from agent $i$ to an eligible target $j$ is their collision-agnostic shortest-path distance from $\Q[i]$; ineligible or unreachable pairs have infinite cost. The minimum-cost one-to-one matching supplies temporary targets to the heuristic and configuration generator. The root computes a full assignment. For a new configuration, only rows belonging to moved agents change, so we reuse the parent's matching and Hungarian dual variables and repair those rows. This incremental update avoids solving the assignment from scratch at every node. Line~\ref{algo:star:goal} accepts a configuration when its agent locations form a one-to-one assignment to eligible targets.

\providecommand{\resultfigdir}{./figures}

\begin{figure*}[t]
    \centering
    \includegraphics[width=0.8\textwidth]
        {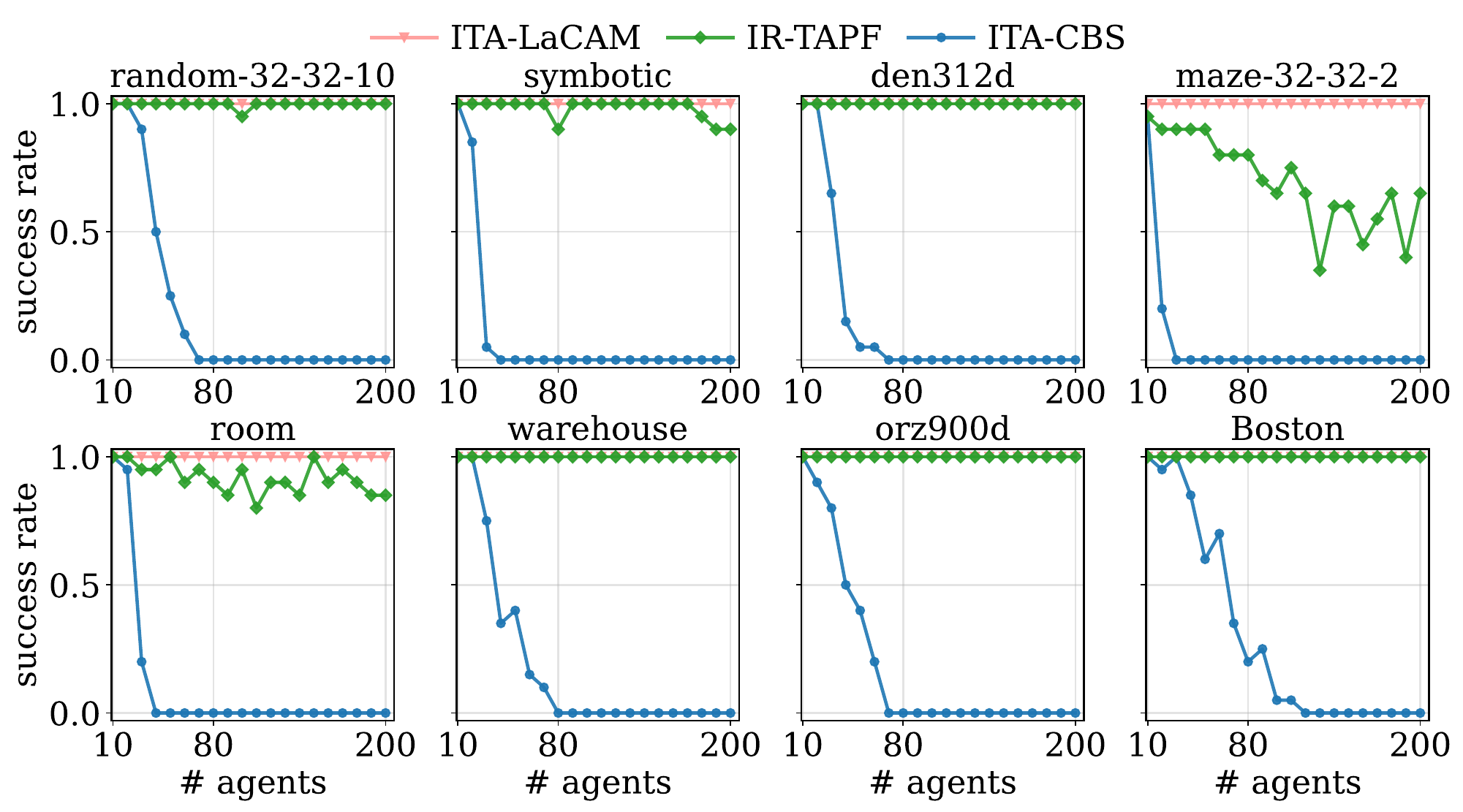}
    \caption{Group Test success rates over 20 instances per point with a 10-second limit. ITA-LaCAM, IR-TAPF, and ITA-CBS solve 100\%, 94.9\%, and 15.9\% of the 3,200 instances, respectively.}
    \label{fig:exp1-success-rates}
\end{figure*}

{
  \begin{algorithm}[ht]
    \caption{Procedure \PIBT{} with Swap}
    \label{algo:pibt-swap}

    \small
    \algrenewcommand\algorithmicif{\textnormal{if}}
    \algrenewcommand\algorithmicthen{\textnormal{then}}
    \algrenewcommand\algorithmicelse{\textnormal{else}}
    \algrenewcommand\algorithmicwhile{\textnormal{while}}
    \algrenewcommand\algorithmicdo{\textnormal{do}}
    \algrenewcommand\algorithmicfor{\textnormal{for}}
    \algrenewcommand\algorithmicend{\textnormal{end}}
    \algrenewcommand\algorithmicreturn{\textnormal{return}}
    \renewcommand{\Continue}{\textnormal{continue}}

    \begin{algorithmic}[1]

      \Input{$i$, $\Q\from$, partial $\Q\to$, and assignment $\tau$}
      \Output{Status \valid{} or \invalid}

      \State $\mathcal{V}_i\gets\neigh(\Q\from[i])\cup\{\Q\from[i]\}$

      \State Sort $\mathcal{V}_i$ in increasing order of $\dist(v,\tau(i))$
      \label{algo:pibt-swap:sort}

      \State $j\gets\funcname{swap\_required\_and\_possible}(i,\mathcal{V}_i[1],\Q\from,\tau)$
      \label{algo:pibt-swap:identify}

      \If{$j\neq\bot$}
        \State Reverse $\mathcal{V}_i$
      \EndIf
      \label{algo:pibt-swap:reverse}

      \For{$v\in\mathcal{V}_i$}

        \If{$v$ conflicts with an assigned move}
          \State \Continue
        \EndIf
        \label{algo:pibt-swap:collision-check}

        \State $\Q\to[i]\gets v$

        \If{$v$ is occupied by an unassigned agent $k$}

          \If{$\PIBT(k,\Q\from,\Q\to,\tau)=\invalid$}
            \State $\Q\to[i]\gets\bot$
            \State \Continue
          \EndIf

        \EndIf
        \label{algo:pibt-swap:call-pibt}

        \If{$v=\mathcal{V}_i[1]\land j\neq\bot\land\Q\to[j]=\bot$}
          \State $\Q\to[j]\gets\Q\from[i]$
        \EndIf
        \label{algo:pibt-swap:pull}

        \State \Return \valid

      \EndFor

      \State $\Q\to[i]\gets\Q\from[i]$
      \State \Return \invalid

    \end{algorithmic}
  \end{algorithm}
}

\providecommand{\resultfigdir}{./figures}

\begin{figure*}[t]
    \centering
    \includegraphics[width=0.8\textwidth]
        {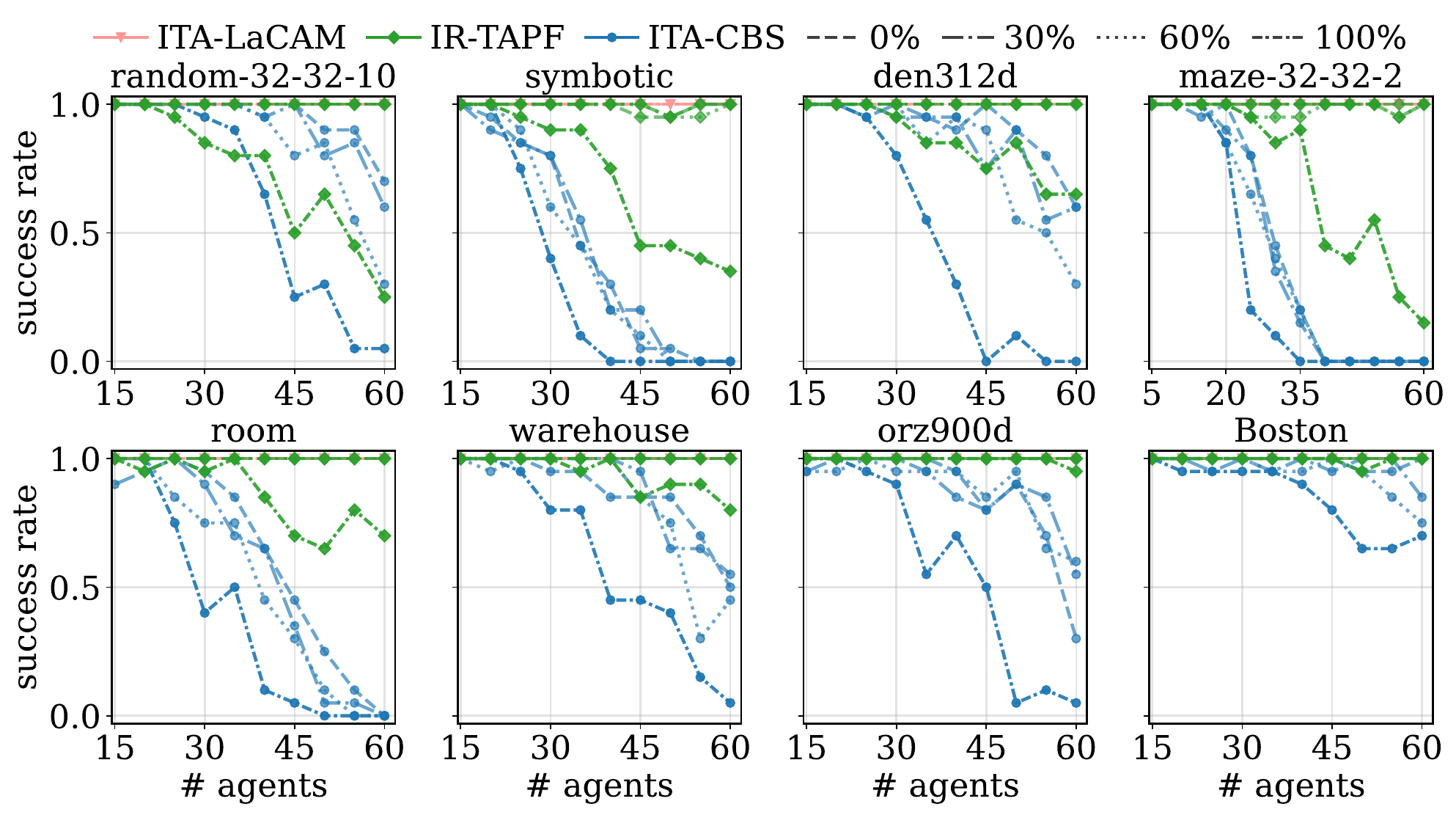}
    \caption{Common Target Test success rates over 20 instances per point with a 10-second limit. ITA-LaCAM, IR-TAPF, and ITA-CBS solve 100\%, 96.0\%, and 67.8\% of 6,560 instances; IR-TAPF's rate drops to 84.6\% when all additional targets are shared.}
    \label{fig:exp2-success-rates}
\end{figure*}

\subsection{PIBT with Swap}

Algorithm~\ref{algo:pibt-swap} details PIBT inside the configuration generator. Given $(\N,\C,\N.\tau)$, the generator sets $\Q\from=\N.\config$, copies the moves fixed by $\C$ into a partial $\Q\to$, and initializes the remaining entries to $\bot$. It calls $\PIBT(i,\Q\from,\Q\to,\N.\tau)$ for each unassigned agent; the calls share $\Q\to$, so each success adds one move. Here, $\neigh(v)$ denotes vertices reachable from $v$ in one move, not the high-level set $\N.\neighbors$. $\mathcal{V}_i$ contains agent $i$'s possible moves and wait action, ordered by shortest-path distance to $\tau(i)$. PIBT examines them in this order and skips a candidate if its vertex is already assigned or it creates an edge collision with an assigned move.

If $v$ is occupied by an unassigned agent $k$, $k$ must move before $i$ can take it. PIBT recursively calls $\PIBT(k,\Q\from,\Q\to,\tau)$ with the same current and partial next configurations. This is priority inheritance: $i$'s request to move is passed to the blocker. If the call succeeds, $i$ keeps $v$; otherwise, it tries the next candidate. If all candidates fail, $i$ waits and returns \invalid{}.

The function $\funcname{swap\_required\_and\_possible}$ handles narrow-corridor livelocks. It checks whether $i$ and a blocking agent $j$ must pass each other and whether a nearby junction provides enough space. It returns $j$ only when both conditions hold. Line~\ref{algo:pibt-swap:reverse} then reverses $i$'s candidate order so it first moves away from its target, and Line~\ref{algo:pibt-swap:pull} assigns $j$ to the vacated location. This creates a multi-step rearrangement rather than a prohibited one-step exchange.


\section{Experiments}

Our main comparison is between ITA-LaCAM and IR-TAPF configured with DBS-Hungarian, two scalable solvers that seek high-quality solutions without an optimality guarantee. We also include ITA-CBS in the success-rate plots and a paired solution-quality comparison as an optimal baseline; it is not the focus of our comparison. ITA-LaCAM and ITA-CBS are implemented in C++, and IR-TAPF is implemented in Rust. Our ITA-LaCAM implementation is based on LaCAM and incorporates optimizations from LaCAM2. Experiments were conducted on Ubuntu 22.04.1 with an Intel Core i9-12900K CPU and 128~GiB of RAM.

We follow the experimental setting of ITA-CBS because preliminary tests using the IR-TAPF setting showed nearly identical runtimes for ITA-LaCAM and IR-TAPF and therefore provided little discriminative power. We replace \texttt{empty-32-32} in the original ITA-CBS benchmark with \texttt{Symbotic}; the resulting eight maps are shown in Figure~\ref{fig:benchmark-maps}.

\subsection{Test Settings}

We use two settings. In the \emph{Group Test}, disjoint groups of five agents share five candidate targets, with 10--200 agents in increments of 10. In the \emph{Common Target Test}, each agent has one private target and 15 additional targets, of which 0, 4, 9 , or 15 are shared, corresponding to common-target ratios of $0\%$, $30\%$, $60\%$, and $100\%$. We test 15--60 agents in incremen ts of five, except on maze, where we test 5--60. Starts and targets are sampled from the same connected component, and every insta nce admits a perfect matching. We use deterministic seeds 0--19 to generate 20 instances for each map, setting, and agent count, a nd all solvers receive the same stored instance and corresponding seed. This yields 3,200 Group Test and 6,560 Common Target Test instances. Each solver has a 10-second limit; a run succeeds if it returns a valid collision-free solution. ITA-LaCAM uses DFS node selection with anytime mode enabled: after finding its first feasible solution, it continues searching within the budget and ret ains its best incumbent according to its internal search cost. IR-TAPF likewise continues iterative refinement and retains its bes t solution. Quality comparisons use only jointly solved instances and report the final-path SOC of the returned incumbents rather than ITA-LaCAM's internal search cost; ITA-CBS reports its proven-optimal solution.

\begin{figure}[t]
    \centering
    \includegraphics[width=0.8\columnwidth]
        {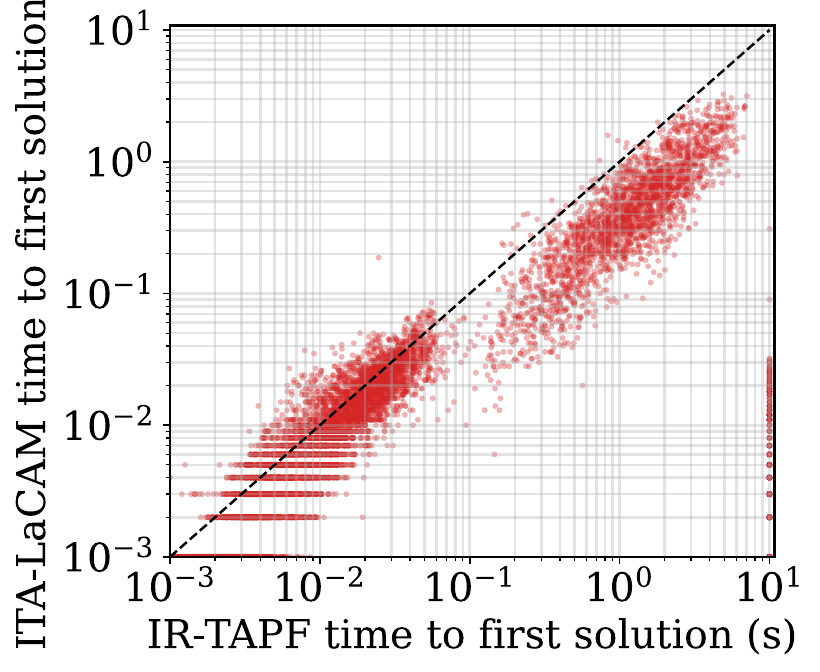}
    \caption{First-solution times on all 9,760 instances; IR-TAPF failures appear on the right boundary ($x=10$~s). Points below the diagonal favor ITA-LaCAM, which is faster on 84.0\% of instances, versus 15.2\% for IR-TAPF and 0.8\% ties.}
    \label{fig:first-solution-time}
\end{figure}

\subsection{Success Rates}

Figures~\ref{fig:exp1-success-rates} and~\ref{fig:exp2-success-rates} compare the two scalable methods with optimal ITA-CBS. ITA-LaCAM solves all 3,200 Group Test and 6,560 Common Target Test instances. In contrast, ITA-CBS solves only 15.9\% of the Group Test, with success falling rapidly as agent counts grow on nearly every map. Its 67.8\% rate in the Common Target Test is higher partly because this test contains at most 60 agents rather than 200. These results illustrate the scalability cost of maintaining optimality. IR-TAPF remains highly scalable overall, solving 94.9\% of the Group Test. Its failures are concentrated on \texttt{maze} and \texttt{room}, while it remains near 100\% on the more open maps. Because Group Test assignments decompose into independent groups of five agents, this pattern suggests that path congestion, rather than global assignment coupling, is the main difficulty in this setting.

The Common Target Test exposes the effect of assignment coupling. IR-TAPF solves 96.0\% overall but only 84.6\% when all additional targets are shared, with the largest drops on \texttt{maze}, \texttt{Symbotic}, and \texttt{random}. ITA-CBS also degrades with more agents and greater overlap, as optimal conflict and assignment resolution becomes expensive. IR-TAPF fixes an assignment until a complete MAPF solution provides feedback, so an unfavorable assignment can consume the time limit. ITA-LaCAM instead repairs its matching at each generated configuration; its 100\% success rate suggests that this shorter feedback loop is most useful when shared targets couple assignment and congestion.

\begin{figure}[t]
    \centering
    \includegraphics[width=0.8\columnwidth]
        {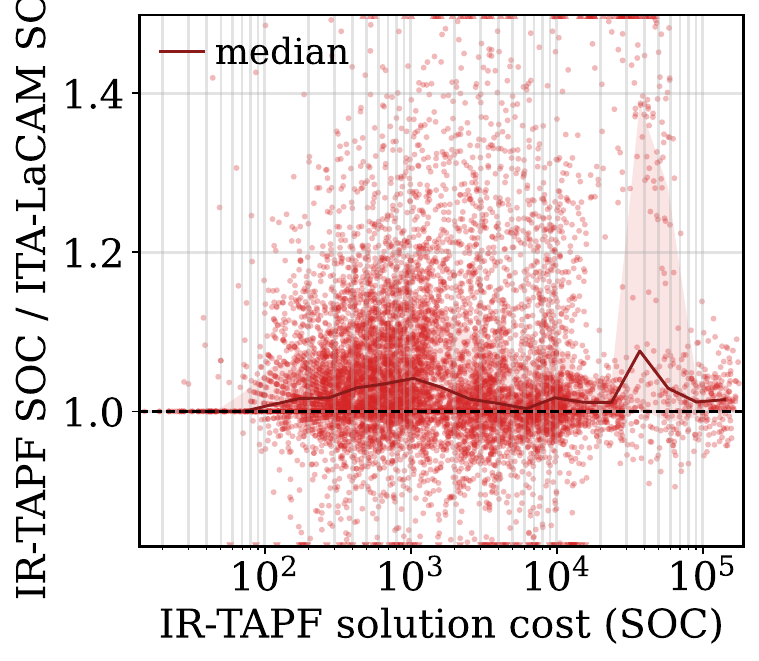}
    \caption{Paired SOC on 9,335 instances (ITA-LaCAM lower: 65.0\%; overall median ratio: 1.020). The curve shows the median ratio among instances with similar IR-TAPF SOC.}
    \label{fig:solution-quality}
\end{figure}

\begin{figure}[t]
    \centering
    \includegraphics[width=0.82\columnwidth]
        {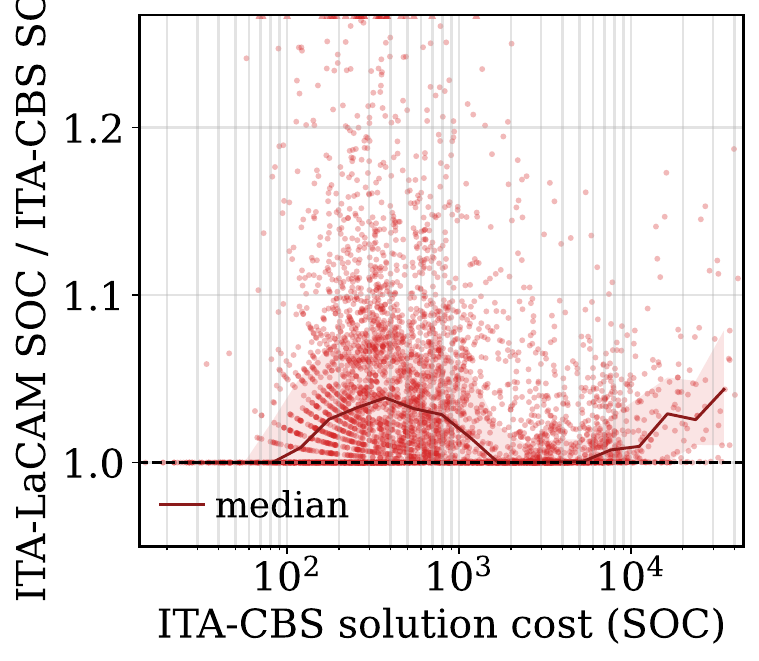}
    \caption{Paired SOC on 4,955 instances (ITA-CBS lower: 67.7\%; overall median ratio: 1.016). The curve shows the median ratio among instances with similar ITA-CBS SOC.}
    \label{fig:solution-quality-itacbs}
\end{figure}

\subsection{Time to the First Solution}

The median first-solution times in Figure~\ref{fig:first-solution-time} are 0.010~s for ITA-LaCAM and 0.014~s for IR-TAPF because both benefit from LaCAM-style path finding. The difference is clearer on harder instances: ITA-LaCAM finds a solution in every case, whereas IR-TAPF misses the limit in 425 cases.

This result reflects where the two methods close the feedback loop between assignment and path finding. IR-TAPF first constructs an assignment and solves the corresponding fixed-target MAPF instance; path feedback can improve the assignment only after that solution is available. ITA-LaCAM uses its current matching directly during successor generation and updates it after every move. It can therefore redirect the configuration search without waiting for a full MAPF solution under a potentially unfavorable assignment. The improvement is thus not merely a small median speedup: the node-wise feedback also reduces the long tail of difficult first-solution times and leaves more of the common time budget for anytime improvement.

\subsection{Solution Quality}

In Figure~\ref{fig:solution-quality}, the binned median remains generally at or above one, indicating that ITA-LaCAM's advantage over IR-TAPF is not driven by a few outliers. Against ITA-CBS in Figure~\ref{fig:solution-quality-itacbs}, 90\% of ITA-LaCAM's paired solutions are within 10.2\% of optimal, although this comparison covers only instances tractable for ITA-CBS.

The two settings reveal the role of assignment coupling. In the Group Test, assignment decomposes into five-agent subproblems; the median ratio is 0.998, and IR-TAPF more often obtains lower SOC (52.3\% versus 42.9\%). Its bottleneck-driven updates are effective when reassignment is local. In the Common Target Test, median ratios increase from 1.000 to 1.170 as target overlap grows, and ITA-LaCAM wins over 90\% of paired cases at $60\%$ and $100\%$ overlap. IR-TAPF is sensitive to its initial assignment~\cite{kumagai2026alternating} and requires a complete MAPF solve before each update. ITA-LaCAM instead revises targets during configuration search, consistent with its growing advantage under stronger coupling.

The remaining gap to ITA-CBS may arise from PIBT and assignment-induced oscillations, both of which introduce redundant motion. Judgelight~\cite{tang2026judgelight} can remove such motion after planning; updating the assignment every few configurations may offer a planning-time alternative between ITA-LaCAM's per-configuration updates and IR-TAPF's per-solution updates.


\section{Conclusion}

We presented ITA-LaCAM, which integrates incremental target assignment into LaCAM-style configuration search. More broadly, our contribution is a shift from treating TAPF as a sequence of fixed-assignment MAPF problems to reasoning over joint configurations and the entire family of MAPF problems induced by feasible assignments within a single search. On 9,760 benchmark instances, ITA-LaCAM solved all within 10 seconds and outperformed IR-TAPF in first-solution time and SOC on 84.0\% and 65.0\% of comparisons, respectively. In the Future, we can explore deep-learning-based guidance adaptable to diverse problem structures and adaptive target-assignment update frequencies between IR-TAPF's per-solution and ITA-LaCAM's per-configuration updates.

\bibliography{strings,references}


\end{document}